\documentclass[letterpaper]{article} 
\usepackage{aaai24}  
\usepackage{times}  
\usepackage{helvet}  
\usepackage{courier}  
\usepackage[hyphens]{url}  
\usepackage{graphicx} 
\usepackage{natbib}  
\usepackage{caption} 
\usepackage{algorithm}
\usepackage{algorithmic}

\usepackage{amsmath}
\usepackage{amssymb}
\usepackage{multirow}
\usepackage{booktabs}
\usepackage{colortbl}
\usepackage{xcolor}
\usepackage{array}
\usepackage{pifont}
\usepackage{bm}
\usepackage[misc]{ifsym}

\def\eg{{\it{e.g.}}}

\def\ie{{\it{i.e.}}}

\definecolor{mycolor}{RGB}{241,240,255}

\usepackage{newfloat}
\usepackage{listings}
\DeclareCaptionStyle{ruled}{labelfont=normalfont,labelsep=colon,strut=off} 
\floatstyle{ruled}
\newfloat{listing}{tb}{lst}{}
\floatname{listing}{Listing}
\title{Towards Compact 3D Representations via Point Feature Enhancement Masked Autoencoders}
\author{
    Yaohua Zha\textsuperscript{\rm 1,2} \quad
    Huizhen Ji\textsuperscript{\rm 1} \quad
    Jinmin Li\textsuperscript{\rm 1} \quad
    Rongsheng Li\textsuperscript{\rm 1} \quad 
    Tao Dai\textsuperscript{\rm 3}\thanks{Corresponding author. (${\text{\Letter}}$ daitao.edu@gmail.com)} \\
    Bin Chen\textsuperscript{\rm 4} \quad
    Zhi Wang\textsuperscript{\rm 1} \quad
    Shu-Tao Xia\textsuperscript{\rm 1,2} \\
}
\affiliations{
    Tsinghua Shenzhen International Graduate School, Tsinghua University\textsuperscript{\rm 1}\\
    Research Center of Artificial Intelligence, Peng Cheng Laboratory\textsuperscript{\rm 2}\\
    College of Computer Science and Software Engineering, Shenzhen University\textsuperscript{\rm 3}\\
    Harbin Institute of Technology, Shenzhen\textsuperscript{\rm 4}\\
    \tt \small chayh21@mails.tsinghua.edu.cn
}

\usepackage{bibentry}

\begin{document}

\maketitle

\begin{abstract}
Learning 3D  representation plays a critical role in masked autoencoder (MAE) based pre-training methods for point cloud, including single-modal and cross-modal based MAE. Specifically, although cross-modal MAE methods learn strong 3D representations via the auxiliary of other modal knowledge, they often suffer from heavy computational burdens and heavily rely on massive cross-modal data pairs that are often unavailable, which hinders their applications in practice. Instead, single-modal methods with solely point clouds as input are preferred in real applications due to their simplicity and efficiency. However, such methods easily suffer from \textit{limited 3D representations} with global random mask input. To learn compact 3D representations, we propose a simple yet effective Point Feature Enhancement Masked Autoencoders (Point-FEMAE), which mainly consists of a global branch and a local branch to capture latent semantic features. Specifically, to learn more compact features, a share-parameter Transformer encoder is introduced to extract point features from the global and local unmasked patches obtained by global random and local block mask strategies, followed by a specific decoder to reconstruct. Meanwhile, to further enhance features in the local branch, we propose a Local Enhancement Module with local patch convolution to perceive fine-grained local context at larger scales. Our method significantly improves the pre-training efficiency compared to cross-modal alternatives, and extensive downstream experiments underscore the state-of-the-art effectiveness, particularly outperforming our baseline (Point-MAE) by \textit{5.16\%}, \textit{5.00\%}, and \textit{5.04\%} in three variants of ScanObjectNN, respectively. Code is available at \url{https://github.com/zyh16143998882/AAAI24-PointFEMAE}. 
\end{abstract}

\section{Introduction}

\begin{table*}[htbp]
  \centering
  \resizebox{\textwidth}{!}{
    \begin{tabular}{llccclllll}
    \toprule
    \multirow{2}[4]{*}{Method} & \multirow{2}[4]{*}{Reference} & \multirow{2}[4]{*}{Input} & \multirow{2}[4]{*}{Cross-Modal Transfer} & \multirow{2}[4]{*}{Masking Strategies} & \multicolumn{3}{c}{Pretrain Efficiency} & \multicolumn{2}{c}{Efficacy} \\
    \cmidrule(lr){6-8}\cmidrule(lr){9-10}   &  &   &  &  & \#Params (M) & GFLOPS & Times (h) & ScanObjectNN & ModelNet40 \\
    \midrule
    \multicolumn{10}{c}{\textit{Single-Modal MAE-based Method}} \\
    \midrule
    Point-MAE & ECCV 2022 & PC    & -     & Global Random & 29.0 (baseline)  & 2.3 (baseline)   & 13 (baseline) & 85.18 (baseline)  & 93.8 (baseline)\\
    Point-M2AE & NeurIPS 2022 & PC    & -     & Multi-Scale Global Random & 15.3 (0.5 $\times$)  & 3.7 (1.6 $\times$)   & 29 (2.2 $\times$)    & 86.43 \textcolor{blue}{($\uparrow$ 1.25)} & 94.0 \textcolor{blue}{($\uparrow$ 0.2)} \\
    \midrule
    \multicolumn{10}{c}{\textit{Cross-Modal MAE-based Method}} \\
    \midrule
    ACT   & ICLR 2023 & PC    & Knowledge Distillation & Global Random & 135.5 (4.7 $\times$) & 31.0 (13.5 $\times$)  & 52 (4.0 $\times$)    & 88.21 \textcolor{blue}{($\uparrow$ 3.03)}   & 93.7 ($\downarrow$ 0.1) \\
    Joint-MAE & IJCAI 2023 & PC \& I & Projection \& 2D Recon. & Global Random & -     & -     & -     & 86.07 \textcolor{blue}{($\uparrow$ 0.89)}   & 94.0 \textcolor{blue}{($\uparrow$ 0.2)} \\
    I2P   & CVPR 2023 & PC \& I & Projection \& 2D Recon. & 2D-Guided & 74.9 (2.6 $\times$)  & 16.8 (7.3 $\times$)  & 64 (4.9 $\times$)    & 90.11 \textcolor{blue}{($\uparrow$ 4.93)}   &94.1 \textcolor{blue}{($\uparrow$ 0.3)} \\
    Recon & ICML 2023 & PC \& I \& L & Contrastive Learning & Global Random & 140.9 (4.9 $\times$) & 20.9 (9.1 $\times$) & 34 (2.6 $\times$)    & \textbf{90.63} \textcolor{blue}{($\uparrow$ 5.45)}  & \textbf{94.5} \textcolor{blue}{($\uparrow$ 0.7)} \\
    \midrule
    \rowcolor{mycolor} Point-FEMAE & Ours     & PC    & -     & Hybrid Global \& Local & 41.5 (1.4 $\times$)  & 5.0 (2.2 $\times$)   & 21 (1.6 $\times$)    & 90.22 \textcolor{blue}{($\uparrow$ 5.04)} & \textbf{94.5} \textcolor{blue}{($\uparrow$ 0.7)} \\
    \bottomrule
    \end{tabular}%
   }
   \caption{Comparison of existing single-modal and cross-modal MAE methods in terms of pre-training efficiency and representational capability. For pre-training efficiency, we evaluate parameters, GFLOPS, and actual pre-training time. For representational capability, we fine-tuned the pre-trained models to evaluate classification accuracy on the ScanObjectNN \cite{scanobjectnn} and ModelNet40 \cite{modelnet}. In the table, PC represents point cloud, I represents images, L represents language, and 2D Recon. refers to 2D image pixels or semantic reconstruction.}
  \label{table1}%
\end{table*}%

Point cloud, as an efficient representation of 3D objects, has been widely used in extensive applications like autonomous driving, robotics, and the metaverse for its rich geometric, shape, and structural details. 
Recently, with the rapid advancements of deep learning-based point cloud understanding \cite{pointnet,dgcnn,xiong2023semantic,gaoimperceptible}, masked autoencoder (MAE) based pre-training methods \cite{pointmae, m2ae, act, i2pmae,recon}, which aim to learn latent 3D representations from vast unlabeled point clouds, have received much attention, and can be categorized into two classes, \ie, {single-modal} \cite{pointmae, m2ae} and {cross-modal} \cite{act, jointmae, i2pmae, recon} methods.

Among them, cross-modal MAE methods, leveraging insights from other modalities, have achieved remarkable performance by acquiring holistic 3D representations.
However, these methods rely heavily on transferring knowledge from massive pair images or texts, which are often unavailable in practice. Specifically, they utilize pre-trained image or language models to extract cross-modal knowledge, along with techniques like projection or knowledge distillation for cross-modal knowledge transfer. Such complex operations require heavy computational cost and thus hinders their applications in practice. 
As shown in Table \ref{table1}, cross-modal methods like Recon \cite{recon} have obtained performance gains by 5\% on ScanObjectNN while requiring $5\times$ pre-training parameters, compared to the single-modal Point-MAE \cite{pointmae}.

For these reasons, single-modal methods with solely point clouds as input are preferred in real applications due to their simplicity and efficiency (Table \ref{table1}). However, existing single-modal methods rely heavily on the global random masked point cloud (shown in Figure \ref{input} (a)) generated by the global random masking strategy to learn 3D representations, which makes the model have robust global shape perception but insufficient local detail representation. As shown in Table \ref{table2}, such single-modal methods can work well on global masked point cloud (GMPC), while failing in local masked point cloud (LMPC), thus resulting in \textit{limited 3D representations} for single-modal MAE models.

To learn compact 3D representations for point cloud, we propose a simple yet highly effective Point Feature Enhancement Masked Autoencoders (Point-FEMAE), which mainly consists of a global branch and a local branch to capture latent global and local features, respectively. Specifically, during the pre-training stage, as illustrated in Figure \ref{framework} (a), we subject a complete point cloud to both global random masking and local block masking to generate globally-biased and locally-biased inputs, respectively. Subsequently, a partially parameter-shared encoder is employed to capture latent global and local features in the global and local branches and rebuild the masked inputs with a branch-independent decoder. Our encoder in both branches shares the same Transformer parameters to ensure comprehensive comprehension of the global points. Furthermore, an additional Local Enhancement Module (LEM) with local patch convolution is introduced within the local branch to perceive fine-grained local context at larger scales. During the fine-tuning phase, as depicted in Figure \ref{framework}(b), owing to the availability of comprehensive global and local information in the complete input point cloud, we employ the encoder from the local branch to learn compact 3D representations of the downstream task point clouds.

Our main contributions are summarized as follows:

\begin{itemize}
    \item We have found that existing single-modal MAE-based point cloud pre-training methods suffer from limited 3D representations, due to the use of a global random masking strategy, which causes biases to the global feature perception while failing to work well on local detail.
    \item We propose a Point Feature Enhancement Masked Autoencoders (Point-FEMAE), which combines global and local mask reconstruction to capture latent enhanced point features. Besides,  a Local Enhancement Module (LEM) is introduced into the encoder to perceive fine-grained local context at larger scales.
    \item Our method significantly improves the pre-training efficiency compared to cross-modal methods. Notably, extensive experiments demonstrate the effectiveness of our method over other MAE-based methods. Particularly, our method significantly outperforms  Point-MAE by 5.16\%, 5.00, and 5.04\% in three variants of ScanObjectNN, respectively.
\end{itemize}

\section{Related Work}

\subsection{Point Cloud Self-supervised Learning}

Self-supervised  Learning (SSL) has achieved remarkable success in many fields such as NLP and computer vision. This approach first applies a pretext task to learn the latent semantic information and then fine-tunes the weights of the model in the target task to achieve higher performance. Existing pretext tasks can be divided into discriminative tasks \cite{becker1992self,wu2018unsupervised,chen2020simple,VLPSU} and generative tasks \cite{mae,lin2021end,data2vec}.
The discriminative approach \cite{pointcontrast} distinguishes different views of the same instance from other instances, and in the point cloud field, PointContrast \cite{pointcontrast} first explores learning 3D representations using contrast learning of features of the same points in different views. CrossPoint \cite{crosspoint} learns point cloud representations within the 3D domain by contrast learning, and then performs further cross-mode contrast learning.
Generation methods \cite{vincent2008extracting,radford2018improving,devlin2018bert,ferles2018denoising,zhang2022survey} typically rely on an autoencoder to learn the latent features of the data by reconstructing the original input. Masked autoencoders (MAE) \cite{mae}, a classical autoencoder that tries to recover the original input from a masked version, which allows the model to learn more robust features, has received a lot of research attention.

\subsection{MAE-based Point Cloud Pre-training}
MAE-based point cloud pre-training methods can be grouped into two categories, \ie, \textit{single-modal} \cite{pointmae, m2ae} and \textit{cross-modal} \cite{act, jointmae, i2pmae, recon} methods. Point-MAE \cite{pointmae} pioneered the use of masked autoencoders for self-supervised pre-training in point clouds. It divides point clouds into patches and employs mini-Point-Net to extract patch embeddings. Then a mask reconstruction was performed with standard transformers and the results were impressive. Afterward, Point-M2AE \cite{m2ae} proposes a multi-scale masking strategy, but still relies on a global random masking strategy at the first scale. Subsequent work mainly focused on using cross-modal knowledge to aid point cloud model learning. For instance, ACT \cite{act} utilized a pre-trained ViT \cite{vit} as a teacher network to guide the learning of the point cloud student network. I2P-MAE \cite{i2pmae} proposed 2D-guided masking and 2D semantic reconstruction to assist point cloud model learning. Recon \cite{recon} learn from both generative modeling teachers and cross-modal contrastive teachers through ensemble distillation. Other MAE-based works \cite{pimae,gdmae,geomae} focus on using scene and LiDAR point clouds for pre-training, specifically for detection tasks. IDPT \cite{idpt} first proposed to introduce prompt tuning in pre-trained point cloud models. Our work focuses on single-modality point cloud pre-training to learn compact 3D representations.

\section{Methodology}

\begin{figure}[t]
    \begin{center}
    \includegraphics[width=\linewidth]{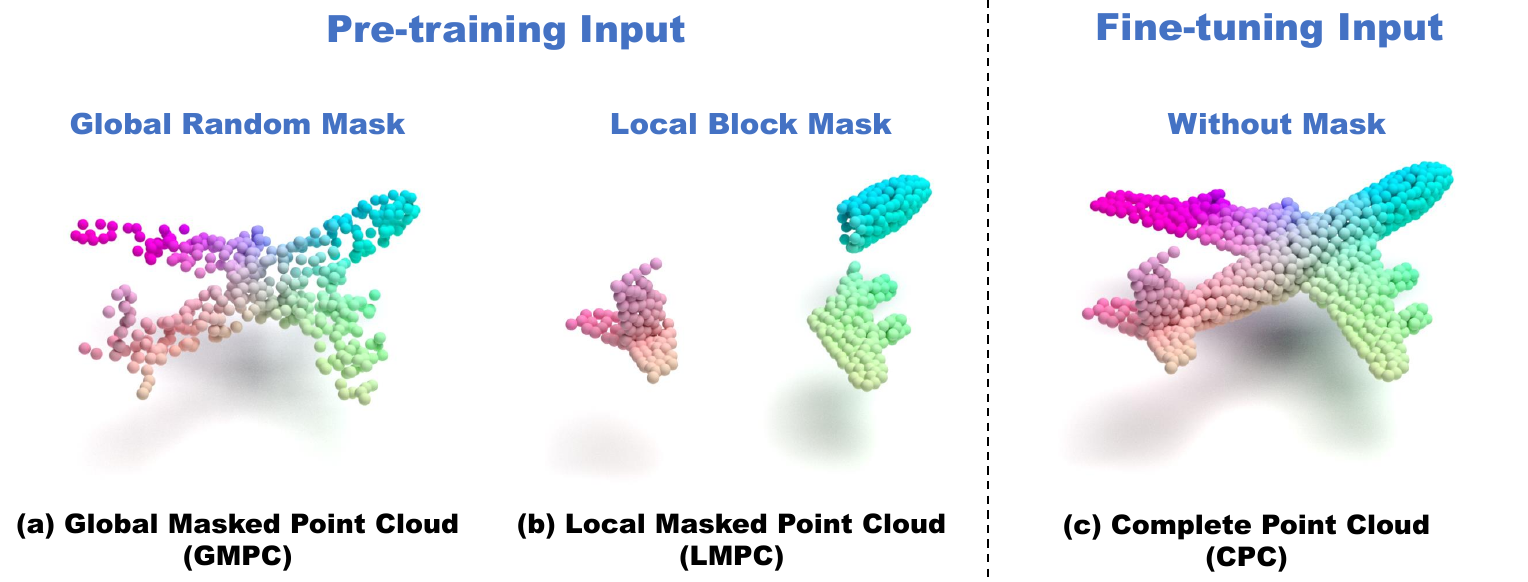}
    \caption{Differences in data distribution between pre-training and fine-tuning. (a) Global Masked Point Cloud (GMPC) input during pre-training with global random masking. (b) Local Masked Point Cloud (LMPC) input during pre-training with local block masking. (c) Complete Point Cloud (CPC) input during downstream fine-tuning.
    }\label{input}
    \end{center}
\end{figure}

\subsection{Observations}

Despite the high efficiency,  existing single-modal MAE-based pre-training pipelines with global/local random mask strategies obtain much worse performance than cross-modal methods (as shown in Table \ref{table1}). It still remains unknown how the random mask strategies affect the single-modal MAE models.
To this end, we first identified a substantial gap in the data distribution between the input data during pre-training and fine-tuning in the context of existing MAE-based methods. During the pre-training stage, conventional masked autoencoders typically employ a global random masking strategy to learn 3D representations, as shown in Figure \ref{input}(a), where a portion of the points is randomly masked. This masking strategy retains the global shape of the point cloud while sacrificing local details. Another strategy of local block masking randomly masks entire point blocks from the complete point cloud at the same ratio, preserving some local details but disrupting global shapes, as shown in Figure \ref{input} (b), which has been demonstrated to yield limited performance \cite{pointbert,pointmae}. However, during the fine-tuning stage, complete point clouds containing full information are often utilized to learn 3D representations, as depicted in Figure \ref{input} (c). 

Our empirical observations suggest that such masked input during the pre-training stage may learn limited 3D representation due to the lack of complete information. Specifically, we employ two straightforward masking strategies: global random mask and local block mask, illustrated in Figure \ref{input} (a) and (b), to dissect the representation efficacy of Point-MAE models pre-trained with these inputs. We assess the models' performance across reconstruction and classification tasks on pertinent test datasets. By introducing point cloud inputs biased toward local details (LMPC) and biased toward global shapes (GMPC) into the model, we gauge its competence in capturing both global and local point representations.

\begin{table}[htbp]
  \centering
  \resizebox{\linewidth}{!}{
    \begin{tabular}{lcccc}
    \toprule
    \multirow{2}[4]{*}{Pre-training Model} & \multicolumn{2}{c}{Reconstruction Chamfer Distance \textcolor{blue}{($\downarrow$)}} & \multicolumn{2}{c}{Classification Accuracy \textcolor{blue}{($\uparrow$)}} \\
    \cmidrule(lr){2-3}\cmidrule(lr){4-5}         & GMPC Input & LMPC Input  & GMPC Input  & LMPC Input  \\
    \midrule
    Point-MAE w/ Global Random Mask & 2.1902  & 2.8538  & 92.77 & 88.98 \\
    Point-MAE w/ Local Block Mask & 2.3533  & 2.4064  & 92.08 & 88.81 \\
    Point-FEMAE (Ours)            & \textbf{2.1880}  & \textbf{2.3941}  & \textbf{93.46} & \textbf{89.33} \\
    \bottomrule
    \end{tabular}%
  }
  \caption{Models with varying mask strategies are assessed using locally-biased LMPC and globally-biased GMPC for classification and reconstruction evaluations. We measure the reconstructed chamfer distance on the ShapeNet test set, lower is better. Additionally, we gauge the classification accuracy during fine-tuning on ScanObjectNN (OBJ-BG), higher is better.}
  \label{table2}%
\end{table}%

The rationale behind this is as follows: for a model utilizing a global random masking strategy, the GMPC inputs are sparsely and randomly spread across the entire object, causing local details to be severely disrupted. Despite this, the overall global shape remains preserved, leading the model to prioritize extracting global features. Conversely, in the case of LMPC inputs, all points are clustered within a few local regions, prompting the model to emphasize learning representations centered on the local surface. Consequently, models exhibiting proficiency in GMPC highlight strong global representation, while those excelling in LMPC underscore potent local representation capabilities.

As illustrated in Table \ref{table2}, the Point-MAE w/ global random masking, demonstrates impressive reconstruction and classification results when tested on GMPC, but its performance is subpar on LMPC. This observation suggests that the model excels in global representation capabilities. Conversely, the Point-MAE w/ local block masking also displays superior performance on GMPC as opposed to LMPC. However, in comparison to global random masking, local block masking encounters a more substantial decline in GMPC performance and a greater enhancement in LMPC performance.

The above observations indicate that existing single modal pre-trained models employing these two straightforward masking strategies lack the ability to excel simultaneously in both LMPC and GMPC, \ie, these models fail to effectively capture both local and global representations. Previous research \cite{pointnet++,dgcnn,deepgcns,pointconv} has demonstrated that models capable of robustly representing both global and local features exhibit higher potential. This insight motivates us to develop a model that learns compact 3D representations by comprehensively exploring global and local information.

\begin{figure*}[t!]
    \begin{center}
    \includegraphics[width=\linewidth]{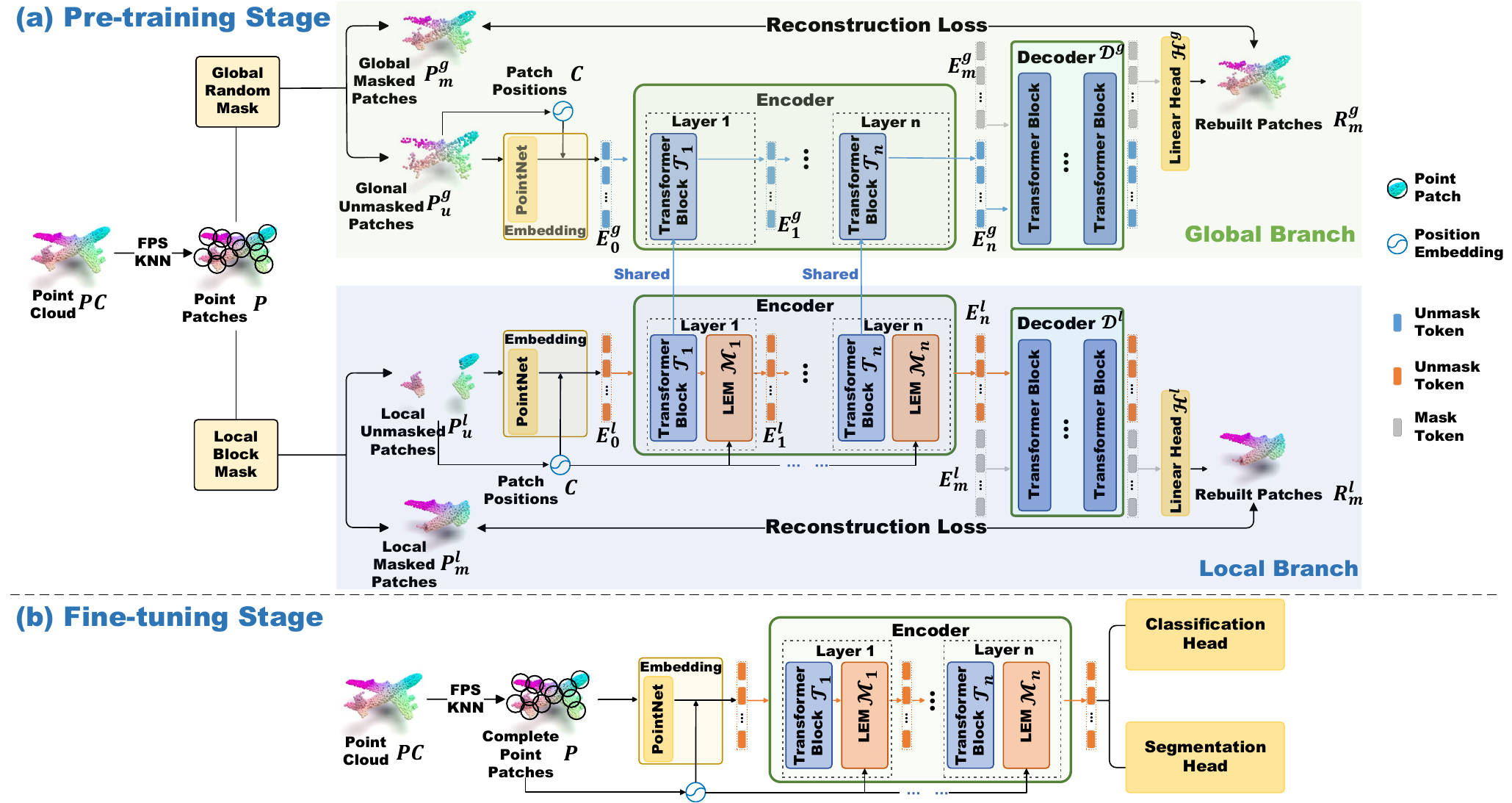}
    \caption{\textbf{The pipeline of our Point-FEMAE.} During the pre-training stage, we perform mask reconstruction in both the global and local branches to learn compact 3D representations. During the fine-tuning stage, we only employ the encoder of the local branch to learn the 3D representation of downstream data.
    }\label{framework}
    \end{center}
\end{figure*}

\subsection{Point Feature Enhancement Mask Autoencoders}

The overall pipeline of our point feature enhancement masked autoencoders (Point-FEMAE) is shown in Figure \ref{framework}. During the pre-training stage, due to the issue of information loss in masked inputs, we performed mask reconstruction in both the global and local branches to learn compact 3D representations. During the fine-tuning stage, owing to the complete input, we only employ the encoder of the local branch to learn the 3D representation of downstream data.

\begin{figure}[t]
    \begin{center}
    \includegraphics[width=\linewidth]{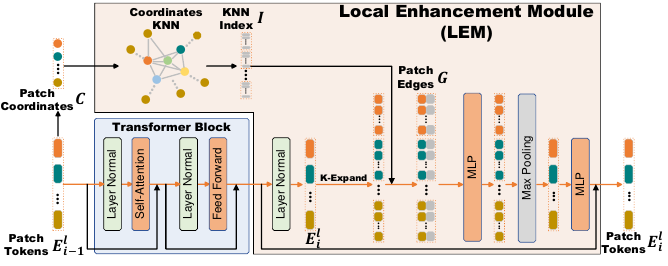}
    \caption{The Encoder Layer's structure, where each layer incorporates a globally oriented Transformer Block and a locally oriented LEM. Within the LEM, information from $k$ nearest neighbor patches is fused based on the patches' coordinates, facilitating a broader scope of local perception.
    }\label{lem}
    \end{center}
\end{figure}

\subsubsection{Masking and Embedding.} Given a point cloud $\bm {PC\in \mathbb{R}^{N\times 3}}$ with $\bm N$ points, we initially divide it into $\bm p$ point patches $\bm {P\in \mathbb{R}^{p \times m \times 3}}$ by farthest point sampling (FPS) and K-Nearest Neighborhood (KNN), with each point patch comprising $\bm m$ local points. Subsequently, in the global branch, we apply global random patch masking to yield unmasked patches $\bm {P_u^g \in \mathbb{R}^{(1-r)p \times m \times 3}} $ and masked patches $\bm {P_m^g \in \mathbb{R}^{rp \times m \times 3}}$, where $\bm r$ denotes the mask ratio. Analogously, within the local branch, we utilize random local block masking to generate $\bm {P_u^l \in \mathbb{R}^{(1-r)p \times m \times 3}}$ and $\bm {P_m^l \in \mathbb{R}^{rp \times m \times 3}}$. 
Finally, $\bm {P_u^g}$ and $\bm {P_u^l}$ are embedded via a light PointNet, and positional encodings are incorporated to derive block tokens $\bm {E_0^g \in \mathbb{R}^{(1-r)p \times C}}$ and $\bm {E_0^l \in \mathbb{R}^{(1-r)p \times C}}$ for the global and local branches, respectively.

\subsubsection{Encoder.} We employ a share-parameter Transformer encode to extract features from the unmasked patches in both the global and local branches. This encoder consists of a series of $\bm n$ encoder layers, each incorporating a standard Transformer block and a Local Enhancement Module (LEM), as depicted in Figure \ref{lem}. The Transformer layer integrates multi-head Self-Attention and a feed-forward network, predominantly focused on perceiving global information. The local enhancement module (LEM), situated after the Transformer Block, is mainly designed to capture local information about the object, during the fine-tuning phase and the local branch of pre-training.

Specifically, for the global branch, during the $\bm i$-th layer forward phase, the feature $\bm {E_n^g}$ only passes through the $\bm i$-th standard Transformer block ${\mathcal T}_i$, allowing the standard Transformer to focus more on the global feature representations. For the local branch, the feature $\bm {E_n^l}$ passes through the $\bm i$-th standard Transformer block and is then fed into the $\bm i$-th Local Enhancement Module ${\mathcal M}_i$, enabling the Local Enhancement Module to focus more on representing local features. Finally, after $\bm n$ layers of forward propagation, the two branches yield the features $\bm {E_n^g}$ and $\bm {E_n^l}$, respectively. 
The forward process of each layer is defined as
\begin{gather}
    \label{eq1}
    {[\bm {E_i^g}; \bm {E_i^l}]}_0 = {[{\mathcal T}_i(\bm {E_{i-1}^g)}; {\mathcal T}_i({\mathcal M}_i(\bm {E_{i-1}^g))}]}_0,
\end{gather}
where $\bm i$ takes values from $\bm 1$ to $\bm n$, and ${[;]}_0$ denotes concatenation along the batch dimension.

\subsubsection{Local Enhancement Module.}
Existing MAE-based methods have exhibited limited local representation, primarily relying on PointNet \cite{pointnet} for extracting patch embeddings to represent limited local contexts. This approach is hindered by two key issues: 1) PointNet inherently lacks localization capabilities, and 2) it struggles to effectively capture localization at broader scales. To tackle these issues, drawing inspiration from Edge-Conv \cite{dgcnn}, we introduce a local patch convolution with coordinate-based nearest neighbors at the patch scale as a dedicated local enhancement module (LEM), to perceive fine-grained local context at larger scales.

Specifically, for each patch token $E^l_{i-1}$, it first undergoes a Transformer Block to yield the current patch tokens $E^l_i$. The patch coordinates $C$ of this patch undergo $K$-Nearest Neighbor (KNN) to obtain the indices $I$ of the $K$ nearest neighboring patches. Through these indices, the relative edges between patches are calculated (\eg, for patches $a$ and $b$ as neighbors, the edge is computed as $E^l_i(a)$ - $E^l_i(b)$). Each patch in $E^l_i$ is then replicated $K$ times, and concatenated with the corresponding edges to form the final edge tensor $G_i$. We apply a single-layer MLP for dimension reduction and use Max pooling to aggregate the $K$ local edges. Lastly, the result goes through another MLP to yield the output tokens $E^l_i$ for the $i$-th layer.

\subsubsection{Decoder.} We employ two distinct decoders, ${\mathcal D}^g$ and ${\mathcal D}^l$, both structured identically. In the local branch, we first concatenate the encoder output $\bm {E^l_n}$ with randomly initialized learnable mask tokens $\bm {E^l_m}$ and direct this composite input into $\bm {D^l}$. Subsequently, we pass the output $\bm {E^l_m}$ through a Linear Head ${\mathcal H}^l$ for coordinate reconstruction, yielding $\bm {R_m^l}$. Finally, we calculate the reconstruction loss between $\bm {R_m^l \in \mathbb{R}^{rp \times m \times 3}}$ and the ground truth $\bm {P_m^l}$. Similar processes are undertaken for the global branch.
Specifically, the forward process of each layer is defined as
\begin{gather}
    \label{eq1}
    \bm {R_m^l} = {\mathcal H}^l({\mathcal D}^l({[\bm {E^l_n} ; \bm {E^l_m}]}_1)[:, \bm {rp}:]) \\
    \label{eq3}
    \bm {R_m^g} = {\mathcal H}^g({\mathcal D}^g({[\bm {E^g_n} ; \bm {E^g_m}]}_1)[:, \bm {rp}:])
\end{gather}
where ${[;]}_1$ denotes concatenation along the token dimension and $[:, \bm {rp}:]$ denotes the last $\bm {rp}$ patch tokens.

\subsubsection{Loss Function.} Following previous works \cite{pointmae}, we use the $\bm {l_2}$ Chamfer Distance \cite{cdloss} ($\mathcal {CD}$) as our reconstruction loss. Our reconstruction target is to recover the coordinates of the local and global branch masked point patches. Our loss function $\mathcal {L}$ is as follows
\begin{gather}
    \label{eq1}
    \mathcal {L}=\mathcal {CD}(\bm {R^g_m},\bm {P_m^g})+ \mathcal {CD}(\bm {R^l_m},\bm {P_m^l})
\end{gather}

\section{Experiments}

\subsection{Pre-training on ShapeNet}

We use ShapeNet \cite{shapenet} as our pre-training dataset, encompassing over 50,000 distinct 3D models spanning 55 prevalent object categories. We extract 1024 points from each 3D model to serve as input for pre-training. The input point cloud is further divided into 64 point patches, with each patch containing 32 points. Table \ref{table1} presents a comparison of our method and other approaches concerning pre-training efficiency and efficacy. 

\subsubsection{Single-Modal.} Compared to the single-modal baseline, Point-MAE \cite{pointmae}, our method shows only slight increases in parameters, GFLOPS, and pre-training time, which are negligible considering the significant performance improvements. In contrast to Point-M2AE \cite{m2ae}, while we possess more parameters and GFLOPS, our pre-training time is notably shorter. This variance arises from Point-M2AE's utilization of a larger input point count and patches (2048 points and 512 patches) in contrast to our utilization of 1024 points and 64 patches.

\subsubsection{Cross-Modal.} In comparison to cross-modal methods, our approach showcases a substantial reduction in parameters (30\%$\sim$60\%), GFLOPS (16\%$\sim$30\%), and pre-training times (30\%$\sim$60\%) due to our simple pipeline and input. Remarkably, while maintaining pre-training efficiency, our method achieves comparable performance to the state-of-the-art cross-modal method, Recon \cite{recon}, underscoring the excellence of our approach.

\subsection{Fine-tuning on Downstream Tasks}

We assess the efficacy of our approach by fine-tuning our pre-trained models on downstream tasks, including classification, few-shot learning, and part segmentation.

\subsubsection{Object Classification.} 

\begin{table*}[htbp]
  \centering
  \resizebox{\textwidth}{!}{
    \begin{tabular}{lccccccccc}
    \toprule
    \multirow{2}[4]{*}{Method} & \multirow{2}[4]{*}{Reference} & \multicolumn{1}{l}{\multirow{2}[4]{*}{\#Params (M)}} & \multicolumn{4}{c}{ScanObjectNN} & \multicolumn{3}{c}{ModelNet40} \\
\cmidrule(lr){4-7}\cmidrule(lr){8-10}        &      &  & Input Data & OBJ-BG & OBJ-ONLY & PB-T50-RS & Input Data & \multicolumn{1}{c}{w/o Vote} & \multicolumn{1}{c}{w/ Vote} \\
    \midrule
    \multicolumn{10}{c}{\textit{Supervised Learning Only}} \\
    \midrule
    PointNet \cite{pointnet} & CVPR 2017 & 3.5   & 1k Points & 73.3  & 79.2  & 68.0    & 1k Points & 89.2  & - \\
    PointNet++ \cite{pointnet++} & NeurIPS 2017 & 1.5   & 1k Points & 82.3  & 84.3  & 77.9  & 1k Points & 90.7  &-  \\
    DGCNN \cite{dgcnn} & TOG 2019 & 1.8   & 1k Points & 82.8  & 86.2  & 78.1  & 1k Points & 92.9  & - \\
    SimpleView \cite{simpleview} & ICML 2021 & -     & 6 Images & -     & -     & 80.5  & 6 Images & 93.9  & - \\
    MVTN \cite{MVTN}  & ICCV 2021 & 11.2  & 20 Images & -     & -     & 82.8  & 12 Images & 93.8  & - \\
    PointMLP \cite{pointmlp}  & ICLR 2022 & 12.6  & 1k Points & -     & -     & 85.2  & 1k Points & 94.1  & 94.5 \\
    SFR \cite{sfr}   & ICASSP 2023 & -     & 20 Images & -     & -     & 87.8  & 12 Images & 93.9  & - \\
    P2P-HorNet \cite{p2p} & NeurIPS 2022 & 195.8 & 40 Images & -     & -     & 89.3  & 40 Images & 94.0  & - \\
    \midrule
    \multicolumn{10}{c}{\textit{Single-Modal Self-Supervised Learning}} \\
    \midrule
    Point-BERT \cite{pointbert} & CVPR 2022 & 22.1  & 1k Points & 87.43 & 88.12 & 83.07 & 1k Points & 92.7  & 93.2 \\
    MaskPoint \cite{maskpoint} & ECCV 2022 & 22.1  & 2k Points & 89.30 & 88.10 & 84.30 & 1k Points & -     & 93.8 \\
    Point-MAE \cite{pointmae} & ECCV 2022 & 22.1  & 2k Points & 90.02 & 88.29 & 85.18 & 1k Points & 93.2  & 93.8 \\
    Point-M2AE  \cite{m2ae} & NeurIPS 2022 & 15.3  & 2k Points & 91.22 & 88.81 & 86.43 & 1k Points &93.4 & 94.0 \\
    \rowcolor{mycolor}  \textbf{Point-FEMAE} & - & 27.4  & 2k Points & \textbf{95.18} & \textbf{93.29} & 90.22 & 1k Points & 94.0 & \textbf{94.5} \\
    \textit{Improvement (baseline: Point-MAE)} & - &  -      & -     & \textcolor{blue}{+5.16}   &  \textcolor{blue}{+5.00}  & \textcolor{blue}{+5.04}  & - & \textcolor{blue}{+0.8}   & \textcolor{blue}{+0.7} \\
    \midrule
    \multicolumn{10}{c}{\textit{Cross-Modal Self-Supervised Learning}} \\
    \midrule
    ACT \cite{act}   & ICLR 2023 & 22.1  & 2k Points & 93.29 & 91.91 & 88.21 & 1k Points & 93.2  & 93.7 \\
    Joint-MAE \cite{jointmae}   & IJCAI 2023 & -  & 2k Points & 90.94 & 88.86 & 86.07 & 1k Points & -  & 94.0 \\
    I2P-MAE \cite{i2pmae} & CVPR 2023 & 15.3  & 2k Points & 94.15 & 91.57 & 90.11 & 1k Points & 93.7  & 94.1 \\
    Recon \cite{recon} & ICML 2023 & 44.3  & 2k Points & \textbf{95.18} & \textbf{93.29} & \textbf{90.63} & 1k Points & \textbf{94.1} & \textbf{94.5} \\
    \bottomrule
    \end{tabular}%
  }
  \caption{Classification accuracy on real-scanned (ScanObjectNN) and synthetic (ModelNet40) point clouds. In ScanObjectNN, we report the overall accuracy (\%) on three variants. In ModelNet40, we report the overall accuracy (\%) for both without and with voting. "\#Params" represents the model's parameters.}
  \label{class}%
\end{table*}%

We initially assess the overall classification accuracy of our pre-trained models on both real-scanned (ScanObjectNN \cite{scanobjectnn}) and synthetic (ModelNet40 \cite{modelnet}) datasets. ScanObjectNN is a prevalent dataset consisting of approximately 15,000 real-world scanned point cloud samples from 15 categories. These objects represent indoor scenes and are often characterized by cluttered backgrounds and occlusions caused by other objects. ModelNet40 is a well-known synthetic point cloud dataset, comprising 12,311 meticulously crafted 3D CAD models distributed across 40 categories.

To ensure a fair comparison, we follow the practices of previous studies \cite{act,recon,i2pmae}. For the ScanObjectNN dataset, we employ data augmentation through simple rotations and report results without voting mechanisms. Additionally, for each input point cloud, we sample 2048 points. Regarding the ModelNet40 dataset, we sample 1024 points for each input point cloud and report overall accuracy for both the without-vote and with-vote configurations and during the fine-tuning phase in ModelNet40, we only update the parameters of our local enhancement modules and the classification head to mitigate overfitting.

As presented in Table \ref{class}, in comparison to baseline Point-MAE, our method showcases substantial enhancements in accuracy across various datasets. Specifically, we observe improvements of 5.16\%, 5.00\%, and 5.04\% on three variants of ScanObjectNN, as well as gains of 0.8\% and 0.7\% on the ModelNet40 (w/o vote and w/ vote respectively). Furthermore, when compared to the leading cross-modal method Recon \cite{recon}, our approach achieves almost equivalent accuracy, while requiring only 62\% of the parameters. These results underscore the unmatched efficiency and efficacy of our pre-trained models, affirming the superiority of our design.

\subsubsection{Few-shot Learning.} 

\begin{table}[htbp]
  \centering
  \resizebox{\linewidth}{!}{
    \begin{tabular}{lcccc}
    \toprule
    \multirow{2}[4]{*}{Method} & \multicolumn{2}{c}{5-way} & \multicolumn{2}{c}{10-way} \\
\cmidrule{2-5}          & 10-shot & 20-shot & 10-shot & 20-shot \\
    \midrule
    \multicolumn{5}{c}{\textit{Supervised Learning Only}} \\
    \midrule
    PointNet \cite{pointnet} & 52.0±3.8 &  57.8±4.9 & 46.6±4.3 & 35.2±4.8 \\
    PointNet-OcCo \cite{wang2021unsupervised} & 89.7±1.9 &   92.4±1.6 & 83.9±1.8 &  89.7±1.5 \\
    PointNet-CrossPoint \cite{crosspoint} & 90.9±4.8 &    93.5±4.4 & 84.6±4.7 &   90.2±2.2 \\
    DGCNN \cite{dgcnn}& 31.6±2.8 & 40.8±4.6 &  19.9±2.1 & 16.9±1.5 \\
    DGCNN-CrossPoint \cite{crosspoint}& 92.5±3.0 &  94.9±2.1 & 83.6±5.3 & 87.9±4.2 \\
    \midrule
    \multicolumn{5}{c}{\textit{Single-Modal Self-Supervised Learning}} \\
    \midrule
    Transformer-OcCo \cite{wang2021unsupervised} & 94.0±3.6 & 95.9±2.3 & 89.4±5.1 & 92.4±4.6 \\
    Point-BERT \cite{pointbert}  & 94.6±3.1 & 96.3±2.7 & 91.0±5.4 & 92.7±5.1 \\
    MaskPoint \cite{maskpoint}  & 95.0±3.7 & 97.2±1.7 & 91.4±4.0 & 93.4±3.5 \\
    Point-MAE \cite{pointmae} & 96.3±2.5 & 97.8±1.8 & 92.6±4.1 & 95.0±3.0 \\
    Point-M2AE \cite{m2ae} & 96.8±1.8 & 98.3±1.4 & 92.3±4.5 & 95.0±3.0 \\
    \rowcolor{mycolor}  Point-FEMAE  & 97.2±1.9 & 98.6±1.3 & \textbf{94.0±3.3} & \textbf{95.8±2.8} \\
    \textit{Improvement (baseline: Point-MAE)} & \textcolor{blue}{+0.9}   &  \textcolor{blue}{+0.8}  & \textcolor{blue}{+1.4}  & \textcolor{blue}{+0.8} \\
    \midrule
    \multicolumn{5}{c}{\textit{Cross-Modal Self-Supervised Learning}} \\
    \midrule
    ACT \cite{act}  & 96.8±2.3 & 98.0±1.4 & 93.3±4.0 & 95.6±2.8 \\
    Joint-MAE \cite{jointmae}  & 96.7±2.2 & 97.9±1.8 & 92.6±3.7 & 95.1±2.6 \\
    I2P-MAE \cite{i2pmae}  & 97.0±1.8 & 98.3±1.3 & 92.6±5.0 & 95.5±3.0 \\
    Recon \cite{recon}  & \textbf{97.3±1.9} & \textbf{98.9±1.2} & 93.3±3.9 & \textbf{95.8±3.0} \\
    \bottomrule
    \end{tabular}%
  }
  \caption{Few-shot learning on ModelNet40. We report the average classification accuracy (\%) with the standard deviation (\%) of 10 independent experiments.}
  \label{fewshot}%
\end{table}%

Following previous works \cite{pointmae, recon}, we conduct few-shot learning experiments on the ModelNet40 \cite{modelnet} dataset using the "$n$-way, $m$-shot" configuration, where $n$ is the number of randomly sampled categories and $m$ is the number of samples in each category. We use the above-mentioned $n \times m$ samples for training, while 20 unseen samples from each category for testing. Following standard protocol, we conducted 10 independent experiments for each setting and reported mean accuracy with standard deviation. 

As indicated in Table \ref{fewshot}, with limited downstream fine-tuning data, our Point-FEMAE exhibits competitive performance among existing single-modal and cross-modal methods, \eg +1.4\% classification accuracy to Point-MAE on the 10-way 10-shot split.

\subsubsection{Part Segmentation.}  We assess the performance of Point-FEMAE in part segmentation using the ShapeNetPart dataset \cite{shapenet}, comprising 16,881 samples across 16 categories. Employing the same experimental settings and segmentation head as Point-MAE and the mean IoU across all categories, i.e., $\mathrm{mIoU}_{c}$ (\%), and the mean IoU across all instances, i.e., $\mathrm{mIoU}_{I}$ (\%) are reported. We did not include the results for Point-M2AE and I2P-MAE due to their utilization of a more intricate segmentation head. 

As shown in Table \ref{seg}, our Point-FEMAE exhibits competitive performance among both existing single-modal and cross-modal methods, \eg +0.7\% $\mathrm{mIoU}_{c}$ to Point-MAE \cite{pointmae} and slightly improvement compared to Recon \cite{recon}. These results demonstrate that our approach exhibits superior performance in tasks such as part segmentation, which demands a more fine-grained understanding of point clouds, demonstrating the superiority of the compact representations learned by our method.

\begin{table}[htbp]
  \centering
  \resizebox{\linewidth}{!}{
    \begin{tabular}{lccc}
    \toprule
    Methods & Reference & $\mathrm{mIoU}_{c}$ & $\mathrm{mIoU}_{I}$ \\
    \midrule
    \multicolumn{4}{c}{\textit{Supervised Learning Only}} \\
    \midrule
    PointNet \cite{pointnet}  & CVPR 2017 & 80.4  & 83.7 \\
    PointNet++ \cite{pointnet++} & NeurIPS 2017 & 81.9  & 85.1 \\
    DGCNN \cite{dgcnn}  & TOG 2019 & 82.3  & 85.2 \\
    PointMLP \cite{pointmlp} & ICLR 2022 & 84.6  & 86.1 \\
    \midrule
    \multicolumn{4}{c}{\textit{Single-Modal Self-Supervised Learning}} \\
    \midrule
    Transformer \cite{attention} & NeurIPS 2017 & 83.4  & 84.7 \\
    Transformer-OcCo \cite{wang2021unsupervised} & ICCV 2021 & 83.4  & 85.1 \\
    Point-BERT \cite{pointbert} & CVPR 2022 & 84.1  & 85.6 \\
    MaskPoint \cite{maskpoint} & ECCV 2022 & 84.4  & 86.0 \\
    Point-MAE \cite{pointmae} & ECCV 2022 & 84.2  & 86.1 \\
    \rowcolor{mycolor} Point-FEMAE & - & \textbf{84.9} & 86.3 \\
    \textit{Improvement (baseline: Point-MAE)} & - & \textcolor{blue}{+0.7}   & \textcolor{blue}{+0.2} \\
    \midrule
    \multicolumn{4}{c}{\textit{Cross-Modal Self-Supervised Learning}} \\
    \midrule
    ACT \cite{act} & ICLR 2023   & 84.7  & 86.1 \\
    Recon \cite{recon} & ICML 2023 & 84.8  & \textbf{86.4} \\
    \bottomrule
    \end{tabular}%
  }
  \caption{Part segmentation results on the ShapeNetPart. The mean IoU across all categories, i.e., $\mathrm{mIoU}_{c}$ (\%), and the mean IoU across all instances, i.e., $\mathrm{mIoU}_{I}$ (\%) are reported.}
  \label{seg}%
\end{table}%

\subsection{Ablation Study}

\subsubsection{Effects of data augmentation, masking strategy, and LEM. } Comparing our fine-tuning with the baseline Point-MAE on ScanObjectNN \cite{scanobjectnn}, our method has three main differences. 1) \textit{Data augmentation}: unlike Point-MAE with scale and translate during fine-tuning, we follow ACT \cite{act} and I2P-MAE \cite{i2pmae} to utilize a simple rotate augmentation. In addition, the data augmentation is the same for all methods in the ModelNet40 dataset. 2) \textit{Masking strategy}: we use a hybrid global and local branch point masking strategy (\eg Hybrid Mask). 3) \textit{Network architecture}: we add our Local Enhancement Module (LEM) after each standard Transformer block in the local branch. We examined the effect of each factor separately. 

We designed four different structures to explore the effects of these factors, as shown in Table \ref{table_ab1}, A1 and A2 use Point-MAE as the baseline, B1 and B2 have a simple hybrid global and local branch mask reconstruction without local enhancement module (LEM), C1 and C2 add our LEM at each layer of the Encoder based on the Point-MAE with global random mask, and D1 and D2 are our Point-FEMAE model. 1 and 2 indicate two different data augmentations.

\begin{table}[htbp]
  \centering
  \resizebox{\linewidth}{!}{
    \begin{tabular}{lcccclll}
    \toprule
    Index & \#Params (M) & Data Aug. & Hybrid Mask & LEM   & OBJ-BG & OBJ-ONLY & PB-T50-RS \\
    \midrule
    A1    & 22.1 & \ding{56}      &  \ding{56}     & \ding{56}      & 90.02 (baseline) & 88.29 (baseline) & 85.18 (baseline) \\
    B1    & 22.1 & \ding{56}      &  \ding{52}     & \ding{56}      & 89.67 \textcolor{gray}{($\downarrow$ 0.35)} & 88.30 \textcolor{blue}{($\uparrow$ 0.01)} & 85.32 \textcolor{blue}{($\uparrow$ 0.14)} \\
    C1    & 27.4 & \ding{56}      &  \ding{56}     & \ding{52}      & 90.36 \textcolor{blue}{($\uparrow$ 0.34)}   & 89.33 \textcolor{blue}{($\uparrow$ 1.04)} & 85.67 \textcolor{blue}{($\uparrow$ 0.49)} \\
    \rowcolor{mycolor} D1    & 27.4 & \ding{56}      &  \ding{52}     & \ding{52}      & 92.77 \textcolor{blue}{($\uparrow$ 2.75)}   & 90.19 \textcolor{blue}{($\uparrow$ 1.90)} & 86.57 \textcolor{blue}{($\uparrow$ 1.39)} \\
    \midrule
    A2    & 22.1 & \ding{52}      &  \ding{56}     & \ding{56}       & 92.94 (baseline) & 92.08 (baseline) & 88.41 (baseline) \\
    B2    & 22.1 & \ding{52}      &  \ding{52}     & \ding{56}       & 92.77 \textcolor{gray}{($\downarrow$ 0.17)} & 91.91 \textcolor{gray}{($\downarrow$ 0.17)} & 88.75 \textcolor{blue}{($\uparrow$ 0.34)} \\
    C2    & 27.4 & \ding{52}      &  \ding{56}     & \ding{52}       & 93.63 \textcolor{blue}{($\uparrow$ 1.19)}   & 92.42 \textcolor{blue}{($\uparrow$ 0.34)} & 89.17 \textcolor{blue}{($\uparrow$ 0.76)} \\
    \rowcolor{mycolor} D2    & 27.4 & \ding{52}      &  \ding{52}     & \ding{52}       & 95.18 \textcolor{blue}{($\uparrow$ 2.24)}   & 93.29 \textcolor{blue}{($\uparrow$ 1.21)} & 90.22 \textcolor{blue}{($\uparrow$ 1.81)} \\
    \bottomrule
    \end{tabular}%
  }
  \caption{Effects of data augmentation, hybrid masking strategy, and LEM on the ScanObjectNN dataset.}
  \label{table_ab1}%
\end{table}%

Table \ref{table_ab1} reports our ablation results, we can discover that: 1) data augmentation leads to a general and noticeable improvement (comparing A1-A2, B1-B2, C1-C2, and D1-D2); 2) simply combining two mask reconstructions can lead to a suboptimal encoder (comparing A1-B1 and A2-B2). 3) introducing LEM to Point-MAE provides a slight improvement (comparing A1-C1 and A2-C2), and this improvement may be due to the introduction of additional parameters, we will discuss this issue in the next subsection. 4) Comparing D1, and D2 with other results, we can discover a significant improvement, which illustrates the superiority of our design, which artfully combines a hybrid global and local branch masking strategy and local enhancement modules.

\subsubsection{Effects of Additional Parameters.} To illustrate whether our improvement is due to more parameters, we introduced the patch-independent MLP and Self-Attention module that focuses on global patches to replace our Local Enhancement Module, respectively, within our masking and reconstruction pipeline for pre-training. We reported their respective fine-tuned results on the ScanObjectNN in Table \ref{table_ab2}.

\begin{table}[htbp]
  \centering
  \resizebox{\linewidth}{!}{
    \begin{tabular}{lcccc}
    \toprule
    Addition Module & \multicolumn{1}{l}{\#Params (M)} & OBJ-BG & OBJ-ONLY & PB-T50-RS \\
    \midrule
    Hybrid Mask w/o LEM & 22.1  &  92.77 &  91.91 & 88.75 \\
    Hybrid Mask w/ 1-layer MLP & 23.9      & 93.63      & 92.43      & 89.14 \\
    Hybrid Mask w/ 3-layer MLPs & 27.4  & 93.80      & 92.43    & 89.17 \\
    Hybrid Mask w/ Self-Attention & 29.2      & 93.98      & 92.60      & 89.42 \\
    Hybrid Mask w/ LEM & 27.4  & 95.18 & 93.29 & 90.22 \\
    \bottomrule
    \end{tabular}%
  }
  \caption{Effects of additional network and parameters.}
  \label{table_ab2}%
\end{table}%

These outcomes demonstrate that incorporating an additional 1-layer MLP exhibits some enhancement when compared to the Hybrid Mask w/o LEM. However, with the escalation of parameters, the model exhibits a limited potential, likely due to the MLP employing shared parameters for individual patch processing, regardless of patch correlations, similar to the Transformer's feed-forward network. Similarly, the additional Self-Attention layer, requiring more parameters, yields a certain improvement, yet it parallels the behavior of the Self-Attention layer within the Transformer, consequently capping potential. These comparisons underscore that the advancement of our approach stems from the excellence of ingeniously combining the strategy of hybrid global and local branch mask reconstruction with the design based on local patch convolution, rather than being driven by additional parameters.

\section{Conclusion}

In this paper, we first compare the pre-training efficiency and efficacy of current single-modal and cross-modal MAE-based point cloud pre-training pipelines and experimentally demonstrate that the limited 3D representation of existing single-modal MAE-based point cloud pre-training methods is due to biases in the existing masking strategies towards global and local representations. To address this issue, we propose to learn compact 3D representations via effective Point Feature Enhancement Masked Autoencoders, which mainly consist of a global branch and local branch to capture latent semantic features. Meanwhile, to further perceive fine-grained local context at larger scales, we propose a Local Enhancement Module with local patch convolution in the local branch. Extensive experiments demonstrate the advancement of our design.

\section{Acknowledgments}

This work is supported in part by the National Key Research and Development Program of China, under Grant No. 2023YFF0905502,  National Natural Science Foundation of China, under Grant (62302309,62171248), Shenzhen Science and Technology Program (Grant No. RCYX20200714114523079, JCYJ20220818101014030, JCYJ20220818101012025), and the PCNL KEY project (PCL2023AS6-1), and Tencent "Rhinoceros Birds" - Scientific Research Foundation for Young Teachers of Shenzhen University.

\bibliography{aaai24}

\clearpage

\section{More Experimental Analysis}

\subsection{Effect of LEM Parameter Count}

\begin{figure}[t!]
    \begin{center}
    \includegraphics[width=\linewidth]{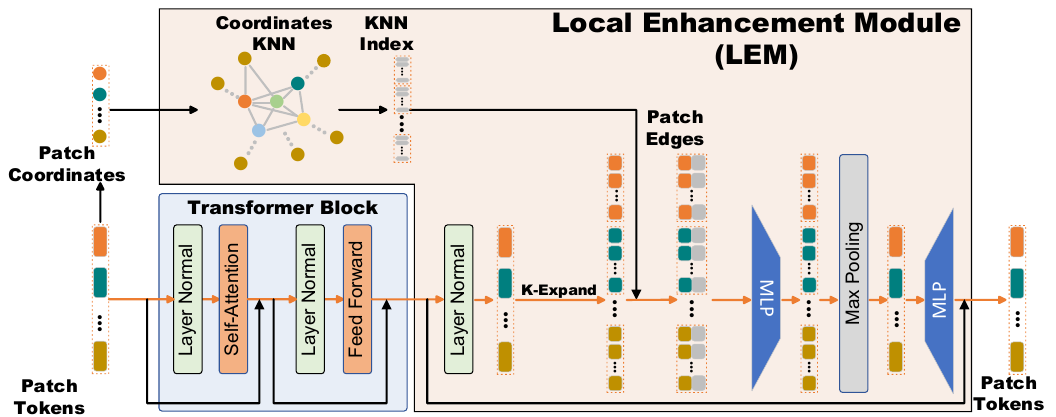}
    \caption{Structure of the Local Enhancement Module (LEM) with Scalable Parameters.
    }\label{lem_s}
    \end{center}
\end{figure}

\begin{table}[htbp]
  \centering
  \resizebox{\linewidth}{!}{
    \begin{tabular}{llll}
    \toprule
    MLP scale & \#Params. (M) & OBJ-BG & OBJ-ONLY \\
    \midrule
    Point-MAE & 22.09 (baseline) & 92.94 (baseline) & 92.08 (baseline)\\
    Point-FEMAE (scale=1/16) & 22.44 \textcolor{blue}{($\uparrow$ 0.35)} & 94.84 \textcolor{blue}{($\uparrow$ 1.90)} & 93.12 \textcolor{blue}{($\uparrow$ 1.04)} \\
    Point-FEMAE (scale=1/8) & 22.78 \textcolor{blue}{($\uparrow$ 0.69)} & 94.84 \textcolor{blue}{($\uparrow$ 1.90)} & \textbf{93.46} \textcolor{blue}{($\uparrow$ 1.38)} \\
    Point-FEMAE (scale=1/4) & 23.44 \textcolor{blue}{($\uparrow$ 1.35)} & 94.84 \textcolor{blue}{($\uparrow$ 1.90)} & 93.29 \textcolor{blue}{($\uparrow$ 1.21)} \\
    Point-FEMAE (scale=1/2) & 24.77 \textcolor{blue}{($\uparrow$ 2.68)} & 94.84 \textcolor{blue}{($\uparrow$ 1.90)} & 93.29 \textcolor{blue}{($\uparrow$ 1.21)} \\
    Point-FEMAE (scale=1) & 27.43 \textcolor{blue}{($\uparrow$ 5.34)} & \textbf{95.18} \textcolor{blue}{($\uparrow$ 2.24)} & 93.29 \textcolor{blue}{($\uparrow$ 1.21)} \\
    \bottomrule
    \end{tabular}%
  \label{table_lem_p}%
  }
  \caption{Effect of LEM Parameter Count.}
\end{table}%

We can alter the parameter count of LEM by scaling the dimensions of the MLP, allowing us to observe the performance of our Point-FEMAE with fewer parameters. Specifically, as illustrated in Figure \ref{lem_s}, we replace the fixed-dimension MLPs in the original LEM with scalable-dimension MLPs and assess the classification accuracy of our Point-FEMAE on the ScanObjectNN \cite{scanobjectnn} dataset for different scales.

Table 1 presents our experimental results. As shown in the table, when progressively reducing the parameters of the Local Enhancement Module (LEM), the model's representation capacity does not show significant degradation. For instance, when scale=1/16, the classification accuracy in both OBJ-BG and OBJ-ONLY only decreases by 0.34\% and 0.17\% respectively, compared to the scale=1 variant. This reduction is negligible considering the substantial improvements relative to the baseline. Additionally, at this point, our method only adds an additional 0.35M parameters, highlighting that the superiority of our approach stems from the design of our Point-FEMAE rather than extra parameters.

\subsection{Effect of Different Architectures and Designs}

\begin{figure*}[t!]
    \begin{center}
    \includegraphics[width=\linewidth]{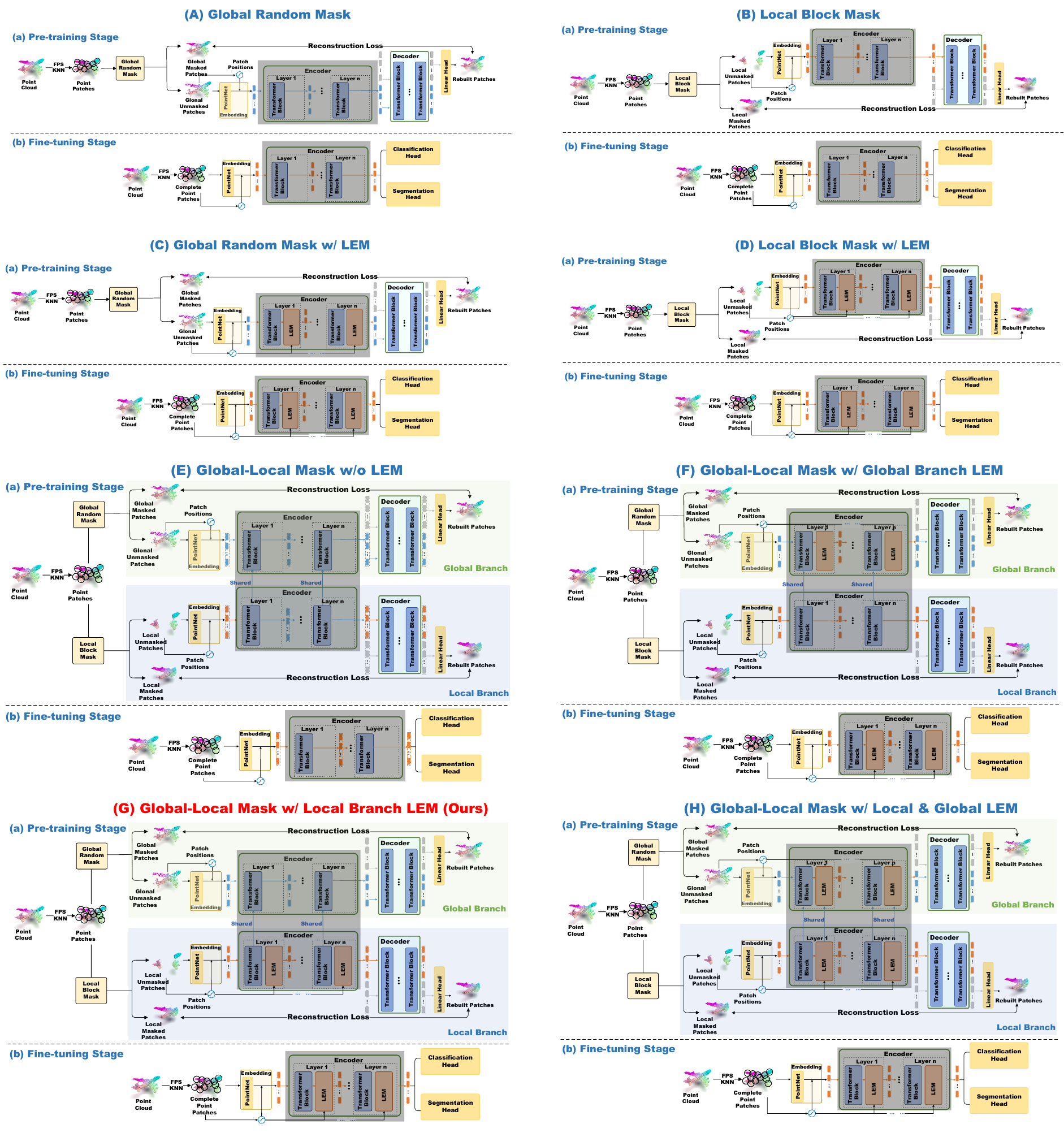}
    \caption{Eight different network architectures based on Point-MAE.
    }\label{arch}
    \end{center}
\end{figure*}

The network structure design and other factors have a certain impact on the experimental results \cite{simpleview,bai2021improving,gudibandetest}. We conducted a more comprehensive analysis to delve into the intricate nuances of various architectures and designs on the single-modal MAE approach. Specifically, we scrutinized eight distinct network architectures, visually depicted in Figure \ref{arch}. 

Among these, Architecture \textbf{A} corresponds to the conventional single-branch Point-MAE \cite{pointmae} utilizing global random masks. Architecture \textbf{B} involves a single-branch Point-MAE approach with the application of local block masks. Architecture \textbf{C} encompasses a single branch Point-MAE method incorporating global random masks and integrated Local Enhancement Modules (LEM). Architecture \textbf{D} represents a single-branch Point-MAE model with local block masks and integrated LEM. Architecture \textbf{E} exclusively employs global-local mask dual branches without integrating LEM. Architecture \textbf{F} adopts global-local mask dual branches and introduces LEM to the global branch. Architecture \textbf{G} is our proposed Point-FEMAE model. Lastly, Architecture \textbf{H} employs global-local mask dual branches and incorporates shared LEMs into both branches.

We report the classification accuracy of these eight architectures on two variants of ScanObjectNN (OBJ-BG and OBJ-ONLY) \cite{scanobjectnn}. All experiments, including those discussed in the main paper, report the highest accuracy achieved across five different random seeds. 

\begin{table}[htbp]
  \centering
  \resizebox{\linewidth}{!}{
    \begin{tabular}{ccll}
    \toprule
    Architecture & \# Params. (M) & OBJ-BG & OBJ-ONLY \\
    \midrule
    A     & 22.1 & 92.94 (baseline) & 92.08 (baseline) \\
    B     & 22.1 & 92.60 \textcolor{gray}{($\downarrow$ 0.34)} & 91.91 \textcolor{gray}{($\downarrow$ 0.17)} \\
    C     & 27.4 & 93.63 \textcolor{blue}{($\uparrow$ 0.69)} & 92.42 \textcolor{blue}{($\uparrow$ 0.34)} \\
    D     & 27.4 & 95.01 \textcolor{blue}{($\uparrow$ 2.07)} & 93.12 \textcolor{blue}{($\uparrow$ 1.21)} \\
    E     & 22.1 & 92.77 \textcolor{gray}{($\downarrow$ 0.17)} & 91.91 \textcolor{gray}{($\downarrow$ 0.17)} \\
    F     & 27.4 & 93.80 \textcolor{blue}{($\uparrow$ 0.86)} & 92.25 \textcolor{blue}{($\uparrow$ 0.17)} \\
    G     & 27.4 & 95.18 \textcolor{blue}{($\uparrow$ 2.24)} & 93.29 \textcolor{blue}{($\uparrow$ 1.12)} \\
    H     & 27.4 & 94.49 \textcolor{blue}{($\uparrow$ 1.55)} & 92.43 \textcolor{blue}{($\uparrow$ 0.35)} \\
    \bottomrule
    \end{tabular}%
  \label{table_supp1}%
}
\caption{Effect of different architectures and designs.}
\end{table}%

By comparing the experimental results in Table 2, we can draw the following conclusions: 1) Within the single-branch designs, the global random masking strategy outperforms the local block masking strategy due to its ability to capture the complete shape of the point cloud (comparing architectures A and B). 2) The inclusion of the Local Enhancement Module (LEM) leads to improved network representation capacity. However, when LEM is incorporated into the global branch, this improvement is significantly limited, as seen in architectures C and F. This limitation arises due to the absence of local details in the global branch, which conflicts with the concept of the Local Enhancement Module. On the contrary, incorporating LEM into the local branch notably enhances performance, as seen in architectures D and G. This is attributed to the local branch effectively preserving the intricate local details of the point cloud. 3) Fully shared Transformers and LEMs (architectures E and H) are suboptimal. This could be attributed to the fact that the different reconstruction targets for local and global features in the shared Encoder might lead to catastrophic forgetting during learning, thereby limiting the model's representation capacity.

\subsection{Effect of Different $K$-Values in LEM}

\begin{figure}[t!]
    \begin{center}
    \includegraphics[width=\linewidth]{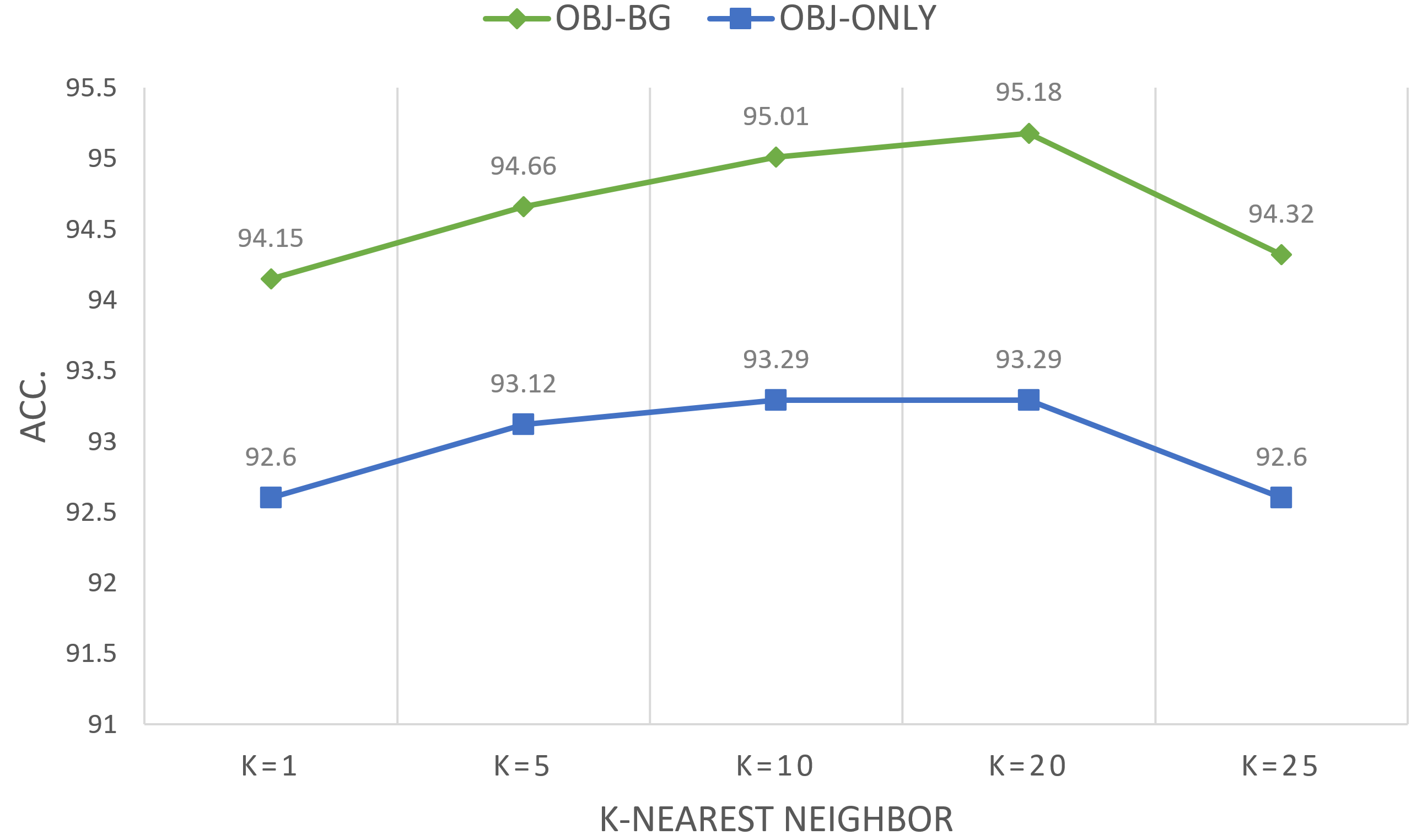}
    \caption{Effect of different $K$-values in LEM.
    }\label{k_nums}
    \end{center}
\end{figure}

The Local Enhancement Module (LEM) enhances local features by employing local convolutions to aggregate information from $K$ nearest neighboring point patches. In this context, we conducted experimental analyses to explore the model's representational capacity when considering varying numbers of $ K$. As illustrated in Figure \ref{k_nums}, our empirical observations suggest that the highest classification accuracy is attained with a $K$ of 20. Additionally, a similar level of capability is achieved when $K$ is set to 10. This implies that the aggregation of patch blocks within the range of 10 to 20 is more effective in capturing fine-grained details within the local structure of the point cloud. For our experimental setup, we selected $K$ = 20.

\subsection{Effect of Different Mask Rates}

\begin{figure}[t!]
    \begin{center}
    \includegraphics[width=\linewidth]{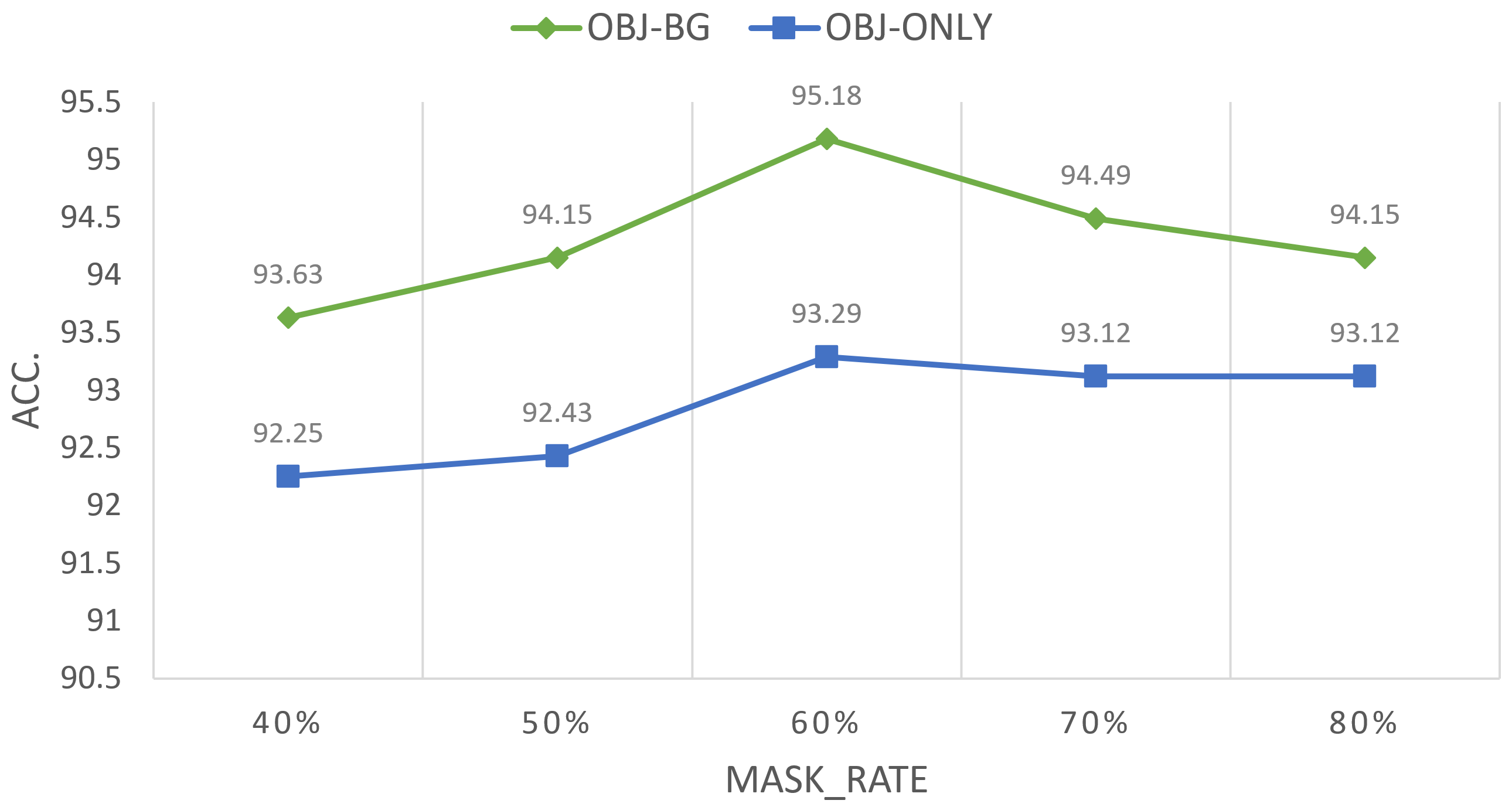}
    \caption{Effect of different mask rates.
    }\label{rate}
    \end{center}
\end{figure}

We conducted experiments to compare the effects of different mask rates during self-supervised pre-training. As shown in Figure \ref{rate}, our empirical observations suggest that using a 60\% mask rate yields relatively better results.

\end{document}